\documentclass{article}

    \usepackage[preprint]{neurips_2025}

\usepackage[utf8]{inputenc} 
\usepackage[T1]{fontenc}    
\usepackage{hyperref}       
\usepackage{url}            
\usepackage{booktabs}       
\usepackage{amsfonts}       
\usepackage{nicefrac}       
\usepackage{microtype}      
\usepackage{xcolor}         
\usepackage{amsmath}
\usepackage{todonotes}
\usepackage[small]{caption}

\newcommand{\concat}[2]{#1 \mathbin{\oplus} #2}
\DeclareMathOperator*{\argmin}{arg\,min}

\title{Can Gradient Descent Simulate Prompting?}

\author{Eric Zhang ~~~~ Leshem Choshen ~~~~ Jacob Andreas \\
MIT CSAIL \\
\texttt{\{zeric,leshem,jda\}@mit.edu} \\
}

\begin{document}

\maketitle

\begin{abstract}
There are two primary ways of incorporating new information into a language model (LM): changing its prompt or changing its parameters, e.g.\ via fine-tuning. Parameter updates incur no long-term storage cost for model changes.
However, for many model updates, prompting is significantly more effective: prompted models can generalize robustly from single examples and draw logical inferences that do not occur under standard fine-tuning. 
Can models be modified so that fine-tuning \emph{does} emulate prompting?
This paper describes a method for meta-training LMs such that gradient updates emulate the effects of conditioning on new information. Our approach uses tools from gradient-based meta-learning but uses an LM's \emph{own prompted predictions} as targets, eliminating the need for ground-truth labels. Subsequent gradient descent training recovers some (and occasionally all) of prompted model performance---showing improvement on the ``reversal curse'' tasks, and answering questions about text passages after a single gradient update. These results suggest that, with appropriate initialization, gradient descent can be surprisingly expressive. Our results suggest new avenues for long-context modeling and offer insight into the generalization capabilities of gradient-based learning.
\end{abstract}

\section{Introduction}
\label{section:intro}
The ability to efficiently and reliably incorporate new information into language models (LMs) is critical for many real-world use cases. Existing approaches typically fall into two categories: context-based methods (e.g., prompting) and parameter-based methods (e.g., full fine-tuning or model editing). While prompting is flexible and effective, it incurs inference-time time and memory costs, and is limited by LMs' context window size. Editing methods, while promising, remain experimental, and their generalization capabilities are poorly understood. Ordinary fine-tuning often fails to produce coherent model updates when applied to individual pieces of new information.

Can we combine the benefits of these different update methods, by building a model that learns from fine-tuning in the same way it learns from context? This would require models to express the behavior modifications caused by context in a static update to the weights, determined by a single gradient. While a large body of work has explored the converse of this question (can in-context learning methods be understood as simulating gradient descent; \citealp{Akyrek2022WhatLA,Oswald2022TransformersLI}, inter alia), it is much less understood whether standard learning algorithms can simulate the more complex predictions that result from conditioning.

We explore this question using a meta-learning objective inspired by MAML \citep{Finn2017ModelAgnosticMF}, searching for a model parameterization that incorporates new information via a gradient step in the same way it does by adding it to context.
In experiments, we show that models meta-trained in this way improve on a variety of tasks, including ``reversal curse'' knowledge edits \citep{Berglund2023TheRC} and passage-based question-answering tasks \citep{Rajpurkar2016SQuAD1Q}.

\begin{figure}
\centering
\includegraphics[width=\linewidth]{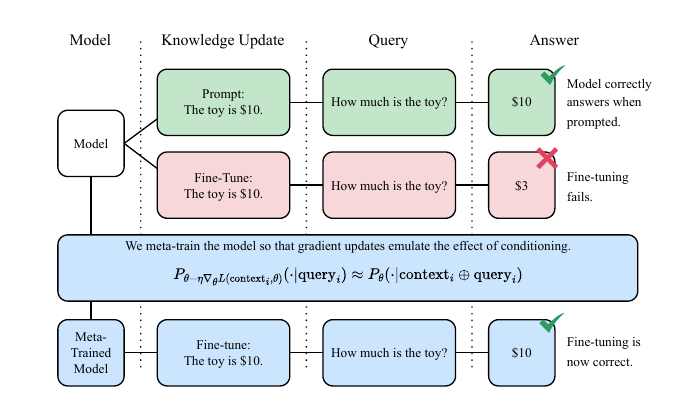}
\caption{\textbf{Meta-training a model to emulate conditioning with fine-tuning.} In the low-sample or 1-sample regime, fine-tuning can fail due to a variety of reasons, including the inability to generalize, the inability to handle reversed phrasing, and difficulty overriding existing priors. On the other hand, models are typically able to use information presented in prompts. We investigate a meta-training procedure to distill the effects of prompting into conditioning, enabling the model to incorporate new information via fine-tuning.}
\label{fig:main}
\end{figure}

\section{Related work}

\paragraph{Prompting} Language models have demonstrated an impressive ability to follow instructions in prompts and perform in-context learning \citep{Brown2020LanguageMA}. As a result, prompting has emerged as a central focus in contemporary research, with numerous studies investigating its behavioral characteristics and underlying mechanisms \citep{Min2022RethinkingTR, Xie2021AnEO, Olsson2022IncontextLA}. Notably, several pieces of recent work have suggested that transformer networks learn in-context via mechanisms similar to gradient descent \citep{Oswald2022TransformersLI, Ahn2023TransformersLT, Dai2023WhyCG}. This is the inverse of what we seek to accomplish: make gradient descent on the models similar to how models learn in-context. Complementary research has explored meta-learning approaches that enhance in-context learning by fine-tuning models on diverse prompt-based tasks \citep{Min2021MetaICLLT, Chen2021MetalearningVL}.

\paragraph{Knowledge Representation} Another line of research has investigated how knowledge is embedded in the parameters \citep{mosbach2024insights}, how to disentangle knowledge in the parameters and context \citep{neeman2023disentqa}, how to remove knowledge from the parameters \citep{zhang2024unforgettablegeneralizationlanguagemodels,Sha2024ForgettingIMA}, and how to embed new information directly into a model's parameters rather than supplying it via the input context, often known as model editing \citep{Zhu2020ModifyingMI, DeCao2021EditingFK, Meng2022LocatingAE, Meng2022MassEditingMI, Liu2022FewShotPF}, which are often closely linked to efforts in model interpretability.

\paragraph{Fine-tuning} Fine-tuning models on multiple examples of a new \emph{task} typically improves downstream performance. However, when applied to update models with single new pieces of knowledge, fine-tuning suffers from many limitations, including poor generalization \citep{Zhu2020ModifyingMI, DeCao2021EditingFK}, inability to override ingrained priors \citep{Onoe2022EntityCB}, and limited ability to model implications of training data (such as information presented with different surface forms \citep{Berglund2023TheRC} or in different languages \citep{ifergan2024beneath}). Moreover, the computation requirements of fine-tuning large models can be prohibitive for those without expensive hardware. Parameter-efficient fine-tuning aims to address these challenges by updating a small subset of parameters. A widely adopted approach is LoRA, which restricts updates to low-rank matrices \citep{Hu2021LoRALA}. For a broader overview of techniques in this area, see the survey by \citet{Lialin2023ScalingDT}.

\paragraph{Gradient-Based Meta-Learning}
Prior to the widespread adoption of large language models, one line of meta-learning research focused on gradient-based techniques like Model-Agnostic Meta-Learning (MAML) \citep{Andrychowicz2016LearningTL, Ravi2016OptimizationAA, Li2017MetaSGDLT, Antoniou2018HowTT, Nichol2018OnFM, Rajeswaran2019MetaLearningWI}. MAML and its extensions employ a bi-level optimization framework using gradients to learn an initialization that enables rapid adaptation to new tasks with minimal data. These methods inspired many adaptations \citep{nichol2018reptile,cho2022fusing} and were widely studied for their effectiveness in few-shot learning scenarios across domains such as computer vision and reinforcement learning \citep{Clavera2018ModelBasedRL, Liu2019TamingME, Sinitsin2020EditableNN}. Recently, some work has been conducted on using gradient-based meta-learning to improve small models' performance on diverse task types \citep{Sinha2024MAMLenLLMMA}.

\paragraph{Context Distillation} Since prompting often outperforms fine-tuning, one approach to inject prompt information into model parameters is to fine-tune a model on generations from the prompted model \citep{Wang2021TowardsZL, Askell2021AGL, Choi2022PromptIP, Akyrek2024DeductiveCT}. Existing work has explored various strategies to improve the effectiveness of this technique, including better sampling strategies and hand-crafted prompts for data augmentation.

\section{Methodology}
\label{section:method}

\subsection{MAML review}

We begin with a review of gradient-based meta-learning \citep[MAML;][]{Finn2017ModelAgnosticMF} as background for our method. Assume we have a dataset of tasks, each of which consists of one demonstration $(x_d, y_d)$ pair and one evaluation pair $(x_e, y_e)$ drawn from the same distribution. We assume that there is one common loss function $L(x, y, \theta) \to \mathbb{R}$ that maps (input, output) pairs and a parameter $\theta$ to a loss. The goal of MAML is to find a model initialization such that fine-tuning the initialization on the demonstration pair makes it effective at predicting the evaluation pair. First, we define the update to the initialization parameters $\theta_0$ that learns $\theta'$:
\begin{equation}
\theta' = \theta_0 - \eta \nabla_{\theta_0} L(x_d, y_d, \theta_0)
\end{equation}
where $\eta$ is some learning rate. We refer to this update as the \textbf{inner loop} of the bi-level optimization problem involved in meta-learning.
Then, the meta-learning loss for this task is the loss of the fine-tuned parameter $\theta'$ on our evaluation pair:
\begin{equation}
L_\mathrm{ML}(x_e, y_e, \theta') = L(x_e, y_e, \theta - \eta \nabla_\theta \mathcal{L}(x_d, y_d, \theta)).
\end{equation}
Given a collection of tasks $\mathcal{T}$, the full meta-learning loss is:
\begin{equation}
\label{eq:maml-outer}
\mathbb{E}_{((x_d, y_d), (x_e, y_e)) \in \mathcal{T}}[L(x_e, y_e, \theta - \eta \nabla_\theta L(x_d, y_d, \theta))]
\end{equation}
We refer to this as the \textbf{outer loop} of the bi-level optimization problem. Optimizing Equation~\ref{eq:maml-outer} yields an initial set of parameters $\theta$ that give good performance on new tasks after a small number of gradient descents.

\subsection{Meta-learning from conditioning}
\label{objective}

The above procedure relies on a set of labeled test examples $x_e$, $y_e$. But suppose we have no such labels, but instead dataset of free-form, natural-language (context, query) pairs $\{(c_i, q_i)\}_N$---for example, (\emph{Bob works at Best Buy.}, \emph{Does Bob work at an electronics store?}).
Finally, suppose we have an off-the-shelf language model with parameters $\theta$. 

It is well established that LMs trained at a sufficiently large scale are capable of answering questions like $q_i$ by \emph{conditioning} on $c_i$. But as noted in Section~\ref{section:intro}, this conditioning operation incurs both storage and processing costs, and if the number of queries an LM must answer is large, we might prefer an LM in which this new information is encoded in parameters rather than input data. Here we describe how a gradient-based meta-learning procedure may be used to accomplish this effect.

Let us denote the conditional distribution of a response given a (context, question) pair:
\begin{equation}
P_\theta(\cdot \mid \concat{c_i}{q_i})
\end{equation}
where $\oplus$ denotes concatenation.
Additionally, let $L(c_i, \theta)$ be the next token prediction loss of the context $c_i$ using the parameters $\theta$. After fine-tuning the parameters $\theta$ on the context $c_i$ with learning rate $\eta$ we have the new parameters:
\begin{equation}
\theta' = \theta - \eta \nabla_\theta L(c_i, \theta).
\end{equation}
This operation corresponds to the inner loop of MAML. We wish to find $\theta$ such that the model behaves the same way after fine-tuning as it does after conditioning. Thus, we want to find parameter $\theta^*$ such that:
\begin{equation}
P_{\theta^*}(\cdot \mid \concat{c_i}{q_i}) \approx P_{\theta^*  - \eta \nabla_{\theta^*} L(c_i, \theta^*)}(\cdot \mid q_i) \quad \forall i.
\label{match}
\end{equation}
At the same time, we want $\theta^*$ to retain broad language modeling capabilities. Let $L_{\text{LM}}(\theta)$ denote a general language modeling loss, and let $\operatorname{D}(P\, \| \, Q)$ be a divergence metric between probability distributions $P$ and $Q$. We can then formulate the optimization objective as:
\begin{align}
\argmin_{\theta^*} \; & \sum_i \operatorname{D}\left(P_{\theta^*}(\cdot \mid \concat{c_i}{q_i}) \,\middle\|\, P_{\theta^* - \eta \nabla_{\theta^*} L(c_i, \theta^*)}(\cdot \mid q_i)\right) 
+  \lambda L_{\text{LM}}(\theta^*)
\end{align}

for some weight $\lambda$. This corresponds to the outer loop of MAML.

\label{simplification}

As conditioning almost always outperforms fine-tuning in our setting, we do not expect to be able to substantially improve $P_{\theta^*}(r_i | \concat{c_i}{q_i})$. Thus, we fix an initial model $\theta_B$ that has been pre-trained as the teacher model. Then, we treat the conditioned distribution using $\theta_B$ as the ``gold'' distribution and fix the left-hand-side of Equation \ref{match}. Also, to avoid the computational expense of sampling multi-token continuations from the conditional distributions when computing the divergence metric, we use the KL divergence of the conditional probabilities of a  greedy decoding from the teacher model. Thus, our method ultimately optimizes:
\begin{align}
\hspace{-1em}
\argmin_{\theta^*} \; & \sum_i \sum_j \operatorname{KL}\left(P_{\theta_B}(\cdot \mid \concat{c_i}{q_i} \oplus \hat{a}_{i,<j}) \,\middle\|\, P_{\theta^* - \eta \nabla_{\theta^*} L(c_i, \theta^*)}(\cdot \mid q_i \oplus \hat{a}_{i,<j})\right) 
+ \lambda L_{\text{LM}}(\theta^*)
\end{align}
where
$\hat{a_i} \sim P_{\theta_B}(\cdot \mid c_i \oplus q_i)$ and $\hat{a}_{i,<j}$ denotes the first $j$ tokens of $\hat{a}$.

\subsection{Meta-learning with ground-truth labels}
\label{upper bound}
To establish an upper bound on performance, our experiments also evaluate standard MAML-style training assuming access to ground-truth answers $a_i$. (This may be viewed as simulating full conditioning accuracy by using gold-responses instead of the conditioning-generated distributions.) Our objective in this case is:
\begin{align}
\argmin_{\theta^*} \; & \sum_i \log P_{\theta^* - \eta \nabla_{\theta^*} L(c_i, \theta^*)}(a_i \mid q_i)
+ \lambda L_{\text{LM}}(\theta^*)
\end{align}
where we use the ground-truth answers $a_i$ as labels.

\subsection{Meta-learning with LoRA}

To reduce the memory required for training, we also experiment with restricting our model to low-rank updates. We examine two possible implementations:

\begin{enumerate}
\item Low-rank update for the model parameters $\theta^*$ in the outer loop, and a step using the same low-rank adapter for the inner step. This experiment evaluates whether it is possible to find a LoRA initialization that is useful for downstream fine-tuning through the meta-learning procedure.
\item Low-rank update on model parameters $\theta^*$ in the outer loop, but a full rank update in the inner step. This experiment evaluates whether we need a high-rank (and potentially complex) update to make the model more capable of fine-tuning, or whether a simpler, low-rank update is enough.
\end{enumerate}

\section{Experiment setup}
\label{section:setup}

\subsection{Data}
\label{section:data}

We focus on four tasks:

\begin{enumerate}
\item \textbf{Character Description} (similar to Reversal Curse \citep{Berglund2023TheRC}): This dataset contains triplets of: a sentence with a description preceding a name, a paraphrased description, and the same name. For instance, a triplet is (``The first person to walk on Mars is Evan Armstrong'', ``The history books record that the first person on Mars is'', ``Evan Armstrong.''). After learning from the first description, the model should be able to complete the paraphrased description with the same name. We use sentences where the description precedes the name because it is easier to evaluate a name completion than a description completion. Due to the limited number of examples in the original Reversal Curse dataset, we generate a new dataset with 5000 training examples and 500 test examples.
\item \textbf{Reversal Curse} \citep{Berglund2023TheRC}: This dataset contains triplets consisting of: a sentence where the name precedes the description, a description, and a name. For instance, a triplet is (``Evan Armstrong is the first person to visit Mars.'', ``The first person to walk on Mars is'', ``Evan Armstrong.''). This is the same as the Character Description dataset, except the sentence the model learns on is reversed. After learning from the first sentence, the model should be able to complete the description with the correct name. Similar to the Character Description task, we generate a new dataset of the same size.

\item \textbf{SQuAD} \citep{Rajpurkar2016SQuAD1Q}: We use the standard SQUAD dataset, which contains triplets of context, question, and answers. After learning from the context, the model should be able to correctly answer the question with an answer. To avoid contamination, we use the splits that come with the dataset. After preprocessing, we have 82031 training records and 10380 test records.

\item \textbf{WikiText} \citep{Merity2016PointerSM}: We transform the WikiText dataset into a next token prediction task. We randomly split the WikiText dataset into triplets. For instance, the text ``It is an adaptation of Strange Case of Dr Jekyll and Mr Hyde, an 1886 novella by Robert Louis Stevenson. The story focuses on the respected London doctor Henry Jekyll,'' can be split into (``It is an adaptation of Strange Case of Dr Jekyll and Mr Hyde, an 1886 novella by Robert Louis Stevenson. The story focuses on'', ``the'', ``respected London doctor Henry Jekyll''). After learning from the the first part of the triplet, the model is expected to complete the second part of the triplet with the third part of the triplet. To avoid contamination, we use the splits that come with the dataset. After preprocessing, we have 343586 train records and 758 test records.
\end{enumerate}

Each dataset is a set of triplets of (context, query, response). There are three ways we can evaluate each triplet:
\begin{enumerate}
\item \textbf{No Context (NC)}: We ignore the context and evaluate the negative log-likelihood (NLL) and accuracy of the response conditioned on the query.
\item \textbf{Prompting (Prompt)}: We evaluate the NLL and accuracy of the response conditioned on the contexts concatenated with the query in the prompt.
\item \textbf{Fine-Tuning (FT)}: We fine-tune on the context and then evaluate the NLL and accuracy of the response conditioned on the query.
\end{enumerate}

As previously noted, it has been widely observed that for most tasks, prompt accuracy > FT accuracy > NC accuracy. Our goal is to increase FT accuracy to match prompt accuracy.

\subsection{Implementation details}

We use gradient descent with a fixed learning rate of $10^{-3}$ for the inner step because we find that it performs well both with and without meta-training. We use Adam \citep{Kingma2014AdamAM} for the outer loop and use held-out data to tune the outer learning rate for each task. Due to the high variance of gradients, we find that aggressive gradient clipping is helpful. We use a batch size of 16, which nearly fills up the VRAM of an 80GB H100 because the second order optimization problem requires VRAM proportional to the product of the model size and batch size to store the adapted parameters after the inner step. Unless otherwise stated, we train for one epoch as we observe that meta-learning starts overfitting substantially after one epoch. We experiment with the Llama 3.2 1B model \citep{Dubey2024TheL3}. Most experiments take several hours on one H100.

\subsection{Fine-tuning preparation}
\label{preparation}

For each listed dataset, we begin by fine-tuning the model on a small subset of dataset in the Prompt and NC configurations. This ensures that the LM has seen the format of the dataset, and isolates the effect of our method from the performance gained from only fine-tuning.

\section{Results}
\label{section:results}

\subsection{Can gradient descent simulate prompting?}
\label{section:results:main}


\begin{figure}
\centering
\includegraphics[width=\textwidth]{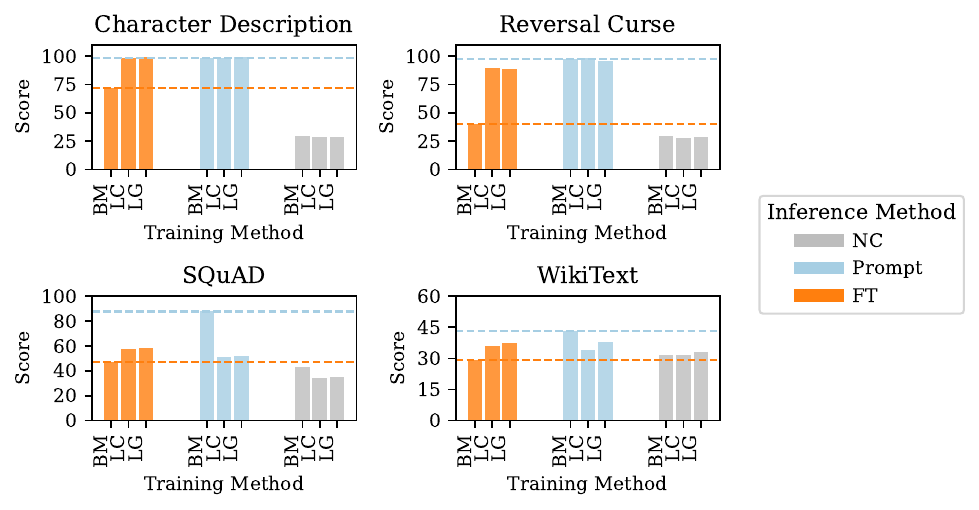}
\caption{\textbf{Performance of models across different training strategies and tasks.} We evaluate the base model (BM), the model after meta-learning from conditioning procedure (LC), and meta-learning from gold labels (LG). We evaluate each task in three different settings. In the No Context (NC) setting, the model is asked to perform the task without any context. This is a lower bound for the performance of our method. In the prompting (Prompt) setting, the model is asked to perform the task with context supplied. This is an upper bound for the performance of our method.  In the Fine-tuning (FT) setting, the model is asked to perform the task after fine-tuning on the appropriate context. Note that the original fine-tuning accuracy of the base model serves as an upper bound for the effectiveness of our method. The using context accuracy of the base model serves as an upper bound for both the LC and the LG procedures. The standard error of all accuracies is less than $\pm 2\%$ over the test set. } 
\label{fig:results}
\end{figure}

Results are shown in Figure~\ref{fig:results}. 
We begin with the general observation that results on learning-from-conditioning are extremely close to those of oracle meta-learning.
For the Character Description task, we observe that our procedure achieves a very high accuracy. This is a simple task, so it serves as a verification that the meta-learning procedure is possible, at least for constrained tasks like these. Interestingly, for the Reversal Curse dataset, the accuracy is lower. The reversed direction is more difficult for the model to learn via a gradient step even with meta-training, though it does improve substantially over the only fine-tuning evaluation of the base model. For the SQuAD dataset, we are only able to recover about a quarter of the performance of the prompted model. Observing the loss curves, we speculate that this is due to a lack of data, as the loss continually decreases until we run out of data to train. For the WikiText dataset, we observe that we are able to recover about half of the performance gap between prompting and a naive fine-tuning. We speculate that the better relative performance of the WikiText dataset is due to the larger dataset size. 


\subsection{Can this effect be achieved with low-rank updates?}
\label{LoRA}

To study the effects of LoRA, we experiment with the SQuAD and WikiText datasets since the Character Description and Reversal Curse datasets are too easy to learn. Since the results of the learning from conditioning and learning from gold responses are very similar as shown in Section \ref{section:results:main}, we present only the results for the learning from gold responses in Table \ref{table:LoRA}, as it is slightly less noisy.

\subsubsection{Can we meta-learn a LoRA initialization that adapts well?}
As displayed in Table \ref{table:LoRA}, we find that the LoRA inner + outer implementation performs very well relative to the full-rank meta-learning and inner step update. The performance is especially better on the SQuAD dataset. We speculate that constraining the rank of the inner step is a strong inductive bias that only a low-rank update is necessary to inject more knowledge \citep{Hu2021LoRALA}. This has a regularizing effect which is especially useful for the SQuAD task, which we have previously observed suffered from a lack of data.


\subsubsection{To improve full-fine-tuning, what rank update do we need?}
We investigate how the meta-learning procedure alters model parameters to improve learning. Specifically, we ask the question of what rank update is necessary to improve downstream fine-tuning.

As shown in Table \ref{table:LoRA}, we find that a rank-1 update achieves comparable performance as a full-rank update. Therefore, this suggests \textit{rank-1 update is enough to make a model exhibit better fine-tuning performance}.

\begin{table}[t]
\centering
\caption{\textbf{Applying LoRA to the gold meta-learning setting.} The 2 columns in the Base Model section give the accuracy using context and fine-tuning accuracies of the base model. The Full Meta section gives the fine-tuning accuracy after performing meta-learning on all of the model parameters. The LoRA Outer Loop section gives the fine-tuning accuracy after meta-training only a rank-1 update on the model parameters and full fine-tuning in the inner step. The LoRA Inner Step section gives the fine-tuning accuracy of an untrained LoRA adapter and a LoRA adapter that has been meta-trained. All results are for the learning from gold responses setting. All reported figures have a standard error within $\pm 2$ over the test set.}
\vspace{0.5em}
\begin{tabular}{lcccccc}
\toprule
& \multicolumn{2}{c}{Base Model} &  \multicolumn{1}{c}{Full Meta} & \multicolumn{1}{c}{LoRA Outer Loop} & \multicolumn{2}{c}{LoRA Inner Step}\\
\cmidrule(lr){2-3} \cmidrule(lr){4-4} \cmidrule(lr){5-5} \cmidrule(lr){6-7}
Dataset & Prompt & FT & FT & FT & Untrained FT & Meta FT\\
\midrule
SQuAD  & 87.7 & 47.3 & 58.6 & 59.4 & 45.1 & 72.0 \\
Wikitext & 43.0 & 29.1 & 37.2 & 37.5 & 32.2 & 38.3\\
\bottomrule
\end{tabular}
\label{table:LoRA}
\end{table}

\subsection{Do the meta-trained models successfully use context?}
When inspecting the meta-trained model's responses to the SQuAD dataset, we display two different ways the model answers correctly in Figure \ref{table:meta_training_combined}.

In the first example, the base model gives an incorrect response both before and after fine-tuning on the relevant context. The meta-trained model also gives an incorrect response before fine-tuning on the context, but successfully responds after fine-tuning on the context. In the second example, the meta-learning process causes the model to automatically respond with the correct answer, even without fine-tuning on the relevant context.

In the second example, meta-learning does not occur because the model can successfully answer the question without taking a gradient on the context. To verify that this is not the main reason behind the improvement, we count the frequency of this behavior in Table \ref{table:meta_training_combined}. We observe that the vast majority of the improvement requires the inner step on the context. This supports the conclusion that the meta-learning procedure is not only teaching the model guess at SQuAD, but to utilize the context through the gradient step.

To further verify that our meta-trained models are actually using the contexts, we also check what happens if we use use a context that is irrelevant. We randomly sample a context from the same distribution, fine-tune the meta-trained model on the context, and then evaluate the model performance. Our evaluations on the SQuAD and WikiText datasets are shown in Table \ref{table:irrelevant}. As shown, the irrelevant contexts harm the performance of the model.

\subsection{Can the meta-trained models retain multiple contexts simultaneously?}
We also investigate if our meta-trained models can retain multiple contexts simultaneously. We study this with the SQuAD dataset because the contexts from SQuAD are non-contradictory, making it possible to retain multiple contexts.

We first evaluate na\"ive continual training of the model from \ref{section:results:main} (which is trained for only a single update) on multiple contexts. We also meta-train two models to explicitly handle multi-context updates by performing a batched update in the inner step instead of an update on a single example. This additionally has the computational advantage of reducing the VRAM usage since this divides the number of adapted parameters we must materialize by the inner batch size.

Our results are shown in Table \ref{table:multiple}. Meta-trained models outperform the base model when fine-tuned on multiple contexts, although there is substantial performance degradation from the single-context version of the method.

\begin{table}[t]
\centering
\caption{\textbf{Learning multiple contexts.} We evaluate the base model, the original meta-trained model, as well as meta-trained models explicitly trained to incorporate multiple contexts at once. The number of contexts a model is meta-trained to handle is specified in parentheses. All evaluations are performed on the SQuAD dataset. We see that the models explicitly trained to handle multiple contexts are better, although there is significant performance degradation. All results are for the learning from gold response setting. All reported results have a standard error less than $\pm 3$ after repeatedly sampling random groups of updates.}
\vspace{0.5em}
\label{table:multiple}
\begin{tabular}{lccccc}
\toprule
Model & 1 Update & 4 Updates & 16 Updates\\
\midrule
Base & 47.3 & 43.6 & 42.6 \\
Meta-Learn (1 ctx trained) & \textbf{57.2} & 46.3 & 39.6 \\
Meta-Learn (4 ctx trained) & 55.1 & \textbf{51.5} & 45.6 \\
Meta-Learn (16 ctx trained)& 55.0 & 47.2 & \textbf{46.6} \\
\bottomrule
\end{tabular}
\end{table}

\begin{table}[t!]
\centering
\caption{\textbf{Cross-dataset meta-learning.} We test if the WikiText meta-trained model transfers to the SQuAD dataset. We perform tests in two settings. In the first sequential setting we perform meta-learning on the WikiText dataset before fine-tuning on the SQuAD dataset. Then we test if the SQuAD dataset also gains meta-learning performance. In the second joint setting, we perform the meta-learning on the WikiText set while fine-tuning on the SQuAD dataset simultaneously. In both settings, we find minor-to-negligible improvements in meta-learning performance via transfer learning. We also test what happens to the WikiText meta-learning performance after both of our procedures, finding that a degree of forgetting occurs in both. All reported figures have a standard error within $\pm 2$ over the test set.}
\vspace{0.5em}
\label{table:cross}
\begin{tabular}{llccc}
\toprule
Method & Evaluation & No Meta-Learning Acc. & In-Domain Acc. & Transfer Acc.\\
\midrule
Sequential & SQuAD FT & 47.3 & 58.6 & 47.8 \\
Joint & SQuAD FT & 47.3 & 58.6 & 48.0 \\
\hline
Sequential & WikiText Retain & 29.1& 37.2 & 34.8 \\
Joint & WikiText Retain & 29.1 & 37.2 & 34.8 \\
\bottomrule
\end{tabular}
\end{table}

\subsection{Does meta-learning transfer across datasets?}

We begin by observing that the WikiText meta-trained model performs poorly on SQuAD. As an example, instead of responding to "Question: The V\&A is looking to open a branded gallery in which city in Scotland?", with "Dundee" like the original Llama model, the meta-trained model begins a full narration: "The answer to that question is..." which is similar to the sentence structure in WikiText. 
We thus conclude with two experiments aimed at understanding: (1) whether a meta-trained model transfers to other datasets and (2) whether fine-tuning a meta-trained model on a downstream task induces catastrophic forgetting of the meta-learning.

In the first experiment, we take the WikiText meta-trained model and fine-tune it on the SQuAD dataset. We are then interested in whether this new model retains the meta-learning performance on the WikiText dataset and whether it is exhibits increased learning performance on the SQuAD dataset.
In the second experiment, we train jointly on the SQuAD fine-tuning task and the WikiText task. We then investigate whether this model exhibits increased learning performance on the SQuAD dataset as well as the performance it achieves on the WikiText meta-learning task. 
For both experiments, we use the exact same subset of the SQuAD dataset that the model was originally fine-tuned on. This prevents the model from being fine-tuned on more of the SQuAD dataset, which will unfairly increase performance.

The results for both of our experiments are present in Table \ref{table:cross}. Surprisingly, we find that the two approaches give similar results: in both cases, it is difficult for the WikiText meta-learning task to generalize to the SQuAD task. 
More surprisingly, we also find that the performance penalty on WikiText after fine-tuning is similar to the performance cost of joint optimization. This suggest that the two types of tasks (fine-tuning vs meta-learning) are different enough that it is difficult to optimize the model for both at once.

\section{Conclusion}

In this paper, we proposed a method for meta-training language models such that gradient updates emulate the effects of conditioning on new information. We find that this method is able to partially bridge the gap between fine-tuning and prompting, enabling models to make some ``prompting-like'' generalizations via gradient-based updates. Surprisingly, we find that a rank-1 update is sufficient to improve a model's ability to fine-tune. Despite these promising results, our method has limitations. Most notably, limited computational resources prevented us from conducting extended meta-training on large, diverse datasets. We hypothesize that scaling up the meta-training process would yield better generalization across tasks and domains. Additionally, future work could investigate methods for better retaining and composing multiple updates, a capability that is essential for robust continual learning.



\bibliographystyle{plainnat}
\bibliography{neurips_2025}

\newpage
\appendix
\section{Technical Appendices and Supplementary Material}

\begin{table}[h]
\centering
\caption{\textbf{Effects and Manner of Meta-Training on Model Performance.} The top panel shows qualitative examples of how meta-training enables better SQuAD answers, either directly or after a gradient step. The bottom panel provides the quantitative distribution of correct responses across validation and training splits, highlighting the impact of context.}
\label{table:meta_training_combined}
\vspace{1em}
\begin{minipage}{\linewidth}
\centering
\textbf{Qualitative Examples of Meta-Training Effects}\\
\vspace{0.2cm}
\begin{tabular}{p{5.5cm}ccp{2cm}c}
\toprule
Question and Correct Answer & Base & Base Step & ML & ML Step\\
\midrule
The V\&A is looking to open a branded gallery in which city in Scotland? Answer: Dundee & Edinburgh & Edinburgh & Edinburgh & Dundee\\
The Rhine forms the border between Austria and what other country? Answer: Switzerland & Germany & Germany & Switzerland & Switzerland\\
\bottomrule
\end{tabular}
\end{minipage}

\vspace{0.5cm}

\begin{minipage}{\linewidth}
\centering
\textbf{Distribution of Correct Responses from Meta-Learned Model}\\
\vspace{0.2cm}
\begin{tabular}{lcccc}
\toprule
Split & Always Incorrect & Always Correct & Correct w/o Context & Only Correct w/ Context\\
\midrule
Val & 55.6\% & 28.0\% & 3.6\% & 12.8\%\\
Train & 29.9\% & 29.7\% & 5.8\% & 34.7\%\\
\bottomrule
\end{tabular}
\end{minipage}
\end{table}

\begin{table}[h]
\centering
\caption{\textbf{Fine-tuning on irrelevant contexts.} We compare the performance of our meta-trained models on the relevant tasks after fine-tuning on the normal context as well as an irrelevant context randomly drawn from the same dataset. As expected, we observe that there is a substantial performance drop. We also display the no context accuracy of a base model for comparison. All reported figures have a standard error within $\pm 2$ over the test set.}
\vspace{1em}
\label{table:irrelevant}
\begin{tabular}{lccc}
\toprule
Dataset & No Context & Irrelevant Context & Normal Context\\
\midrule
SQuAD &35.2 & 29.7 & 58.6\\
WikiText & 33.1 & .010 &37.2\\
\bottomrule
\end{tabular}
\end{table}

\end{document}